\DeclareUnicodeCharacter{FB01}{fi}
\documentclass[runningheads]{llncs}
\usepackage{graphicx}
\usepackage{multirow}
\usepackage{amsmath}
\usepackage{amssymb}
\usepackage{wrapfig}
\usepackage{marvosym}
\usepackage{hyperref}

\begin{document}
\title{SIN: Superpixel Interpolation Network}
\author{Qing Yuan\inst{1}
\and
Songfeng Lu\inst{2,3}\textsuperscript{(\Letter)} \and
Yan Huang\inst{1}
\and
Wuxin Sha\inst{1}}
\authorrunning{Q. Yuan Author et al.}

\institute{School of Computer Science and Technology, Huazhong university of Science and Technology, Wuhan, China
\and
School of Cyber Science \& Engineering, Huazhong university of Science and Technology, Wuhan, China
\and 
Shenzhen Huazhong University of Science and Technology Research Institute, Shenzhen, China
\\	
\email{\{yuanqing,lusongfeng,m201372777,d201980975\}@hust.edu.cn}}
\maketitle              % typeset the header of the contribution
\begin{abstract}
Superpixels have been widely used in computer vision tasks due to their representational and computational efficiency.
Meanwhile, deep learning and end-to-end framework have made great progress in various fields including computer vision.
However, existing superpixel algorithms cannot be integrated into subsequent tasks in an end-to-end way. Traditional algorithms and deep learning-based algorithms are two main streams in superpixel segmentation. The former is non-differentiable and the latter needs a non-differentiable post-processing step to enforce connectivity, which constraints the integration of superpixels and downstream tasks.
In this paper, we propose a deep learning-based superpixel segmentation algorithm SIN which can be integrated with downstream tasks in an end-to-end way. Owing to some downstream tasks such as visual tracking require real-time speed, the speed of generating superpixels is also important.
To remove the post-processing step, our algorithm enforces spatial connectivity from the start.
Superpixels are initialized by sampled pixels and other pixels are assigned to superpixels through multiple updating steps. Each step consists of a horizontal and a vertical interpolation, which is the key to enforcing spatial connectivity. Multi-layer outputs of a fully convolutional network are utilized to predict association scores for interpolations. 
Experimental results show that our approach runs at about 80fps and performs favorably against state-of-the-art methods. Furthermore, we design a simple but effective loss function which reduces much training time.
The improvements of superpixel-based tasks demonstrate the effectiveness of our algorithm. We hope SIN will be integrated into downstream tasks in an end-to-end way and benefit the superpixel-based community. Code is available at: \href{https://github.com/yuanqqq/SIN}{https://github.com/yuanqqq/SIN}.

\end{abstract}
\section{Introduction}
Superpixels are small clusters of pixels that have similar intrinsic properties. Superpixels provide a perceptually meaningful representation of image data and reduce the number of image primitives for subsequent tasks. Owing to their representational and computational efficiency, superpixels are widely applied to computer vision tasks such as object detection \cite{object1,object2}, saliency detection \cite{saliency1,saliency2,SO}, semantic segmentation \cite{semantic1,semantic2,semantic3} and visual tracking \cite{track1,track2}. 

In common, superpixel-based tasks first generate superpixels of input images.
Afterwards, features of superpixels are extracted and fed into subsequent steps.
Since most superpixel algorithms cannot ensure spatial connectivity directly, we need to enforce spatial connectivity through a post-processing step before extracting superpixel features.
Recently, deep neural networks and end-to-end framework have been widely adopted in computer vision owing to their effectiveness.
However, existing superpixel segmentation algorithms cannot be combined with downstream tasks in an end-to-end way, which constrains the application of superpixels and the performance of superpixel-based tasks. We will demonstrate the limitations of existing superpixel segmentation algorithms in the following.

Existing superpixel segmentation algorithms can be divided into traditional and deep learning-based branches.
Traditional superpixel segmentation algorithms \cite{NC,FH,ERS,SEEDS,SLIC,SNIC} mainly rely on hand-crafted features. They are not trainable and cannot be integrated to subsequent deep learning methods in an end-to-end way obviously. Not to mention that most traditional algorithms run at a low speed, which affects the speed of downstream tasks heavily.
While few attempts have been made ~\cite{SEAL,SSN,superpixel_fcn}, utilizing deep networks to extract superpixels remains challenging. \cite{SEAL,SSN} use a deep network to extract pixel features, followed by a superpixel segmentation module. FCN~\cite{superpixel_fcn} proposes a network to directly generate superpixels and enforce connectivity as a post-processing step.
All these methods need a post-processing step to handle orphan pixels and the step is non-differentiable. The post-processing step hinders existing deep learning-based algorithms to be combined with superpixel-based tasks in an end-to-end way. In fact, most traditional algorithms also need post-process to enforce spatial connectivity.

In this paper, we aim to propose a superpixel segmentation algorithm which can be integrated into downstream tasks in an end-to-end way.
The speed of generating superpixels is also very important, because some downstream tasks such as visual tracking require real-time speed.
Since the post-processing step is the main obstacle of existing deep learning-based methods, we enforce spatial connectivity from the start to remove the step.
Without the post-processing step, not only the algorithm becomes a whole trainable network, but also the speed is faster.
Our initial superpixels are initialized with sampled pixels and remaining pixels are assigned to superpixels through multiple similar steps. Each step consists of a horizontal and a vertical interpolation.
According to current pixel-superpixel map and association scores, the interpolations assign partial pixels to superpixels.
The pixel-superpixel map represents the map between pixels and superpixels, and the association scores are predicted by the multi-layer outputs of a fully convolutional network. 
The rule of interpolations is the key to enforcing spatial connectivity and we will prove it in Section~\ref{section:connectivity}.
Furthermore, we design a simple but effective loss function that can reduce training time and fully utilize segmentation labels.

Extensive experiments have been conducted to evaluate SIN.
Our method is the fastest compared to existing deep learning-based algorithms(running at about 80fps), which means it satisfies the instantaneity of downstream tasks.
For superpixel segmentation, experimental results on public benchmarks such as BSDS500~\cite{BSD} and NYUv2~\cite{NYU} demonstrate that our method performs favorably against the state-of-the-art in a variety of metrics. 
For semantic segmentation and saliency object detection, we replace superpixels in the original BI~\cite{BI} and SO~\cite{SO} with ours. The results on PascalVOC 2012 test set~\cite{VOC} and ECSSD dataset~\cite{ecssd} show that SIN superpixels benefit these downstream vision tasks.

In summary, the main contributions of this paper are:
\begin{itemize}
   \item We propose a superpixel segmentation network which can be integrated into downstream tasks in an end-to-end way, which does not need post-processing to handle orphan pixels.
   Our algorithm enforces spatial connectivity from the first instead of using a non-differentiable post-processing step.
   To the best of our knowledge, we are the first to develop a deep learning-based method to be integrated into superpixel-based tasks in an end-to-end way. 
   \item We analyze the runtime of deep learning-based superpixel algorithms and our model has the fastest speed.
   When utilizing our SIN superpixels in subsequent tasks, the instantaneity will not be destroyed. Extensive experiments show that our method performs well in superpixel segmentation especially in generating more compact superpixels.
   \item We design a simple but effective loss function that fully utilizes the segmentation label. The loss function is computational efficiency and shortens plenty of training time.  
\end{itemize}

\section{Related Work}
\subsection{Traditional Superpixel Segmentation}
Traditional superpixel segmentation  algorithms can be roughly categorized as graph-based and clustering-based algorithms.
Graph-based algorithms treat image pixels as graph nodes and pixel affinities as graph edges.
Usually, superpixel segmentation problems are solved by graph-partitioning.
~\cite{NC} applies the Normalized Cuts algorithm to produce the superpixel map.
FH~\cite{FH} defines an adaptive segmentation criterion to capture global image properties.
ERS~\cite{ERS} proposes an objective function for superpixel segmentation, which consists of the entropy rate and the balancing term.

Clustering-based algorithms utilize clustering methods such as $k$-means for superpixel segmentation.
SEEDS~\cite{SEEDS} starts from an initial superpixel partitioning and continuously exchanges pixels on the boundaries between neighboring superpixels.
SLIC~\cite{SLIC} adopts a $k$-means clustering approach to generate superpixels based on a 5-dimensional positional and $Lab$ color features.
Owing to its simplicity and high performance, there are many variants \cite{LSC,Manifold,SNIC} of SLIC.
LSC~\cite{LSC} projects the 5-dimensional features to a 10-dimensional space and performs weighted $k$-means in the projected space.
Manifold-SLIC~\cite{Manifold} maps the image to 2-dimensional manifold feature space for superpixel clustering.
SNIC~\cite{SNIC} proposes a non-iterative scheme for superpixel segmentation. Traditional superpixel algorithms are mainly based on hand-crafted features, which often fail to preserve weak object boundaries.
% 1. low speed 2. perform not so well 3. cannot combine
Most traditional algorithms are computed on CPU, so it is hard to achieve real-time speed.
What's more, we cannot integrate traditional methods into subsequent tasks in an end-to-end way because they are non-differentiable.

\subsection{Superpixel Segmentation using DNN}

Recently, some researchers have focused on integrating deep networks into superpixel segmentation algorithms \cite{SEAL,SSN,superpixel_fcn}.
\cite{SEAL,SSN} use a deep network to extract pixel features, which are then fed to a superpixel segmentation module.
SEAL~\cite{SEAL} develops the Pixel Affinity Net for affinity prediction and defines a new loss function which takes the segmentation error into account.
These affinities are then passed to a graph-based algorithm to generate superpixels. 
To form an end-to-end trainable network, SSN~\cite{SSN} turns SLIC into a differentiable algorithm by relaxing the nearest neighbors' constraints.
FCN~\cite{superpixel_fcn} combines feature extraction and superpixel segmentation into a single step.
The proposed method employs a fully convolutional network to predict association scores between image pixels and regular grid cells. 
When utilizing superpixels generated by existing deep learning-based methods, a post-processing step is needed to handle orphan pixels.
The step is not trainable and can only be computed on CPU, so existing deep learning-based methods cannot be integrated into downstream tasks in an end-to-end approach.

\subsection{Spatial Connectivity}
Most superpixel algorithms \cite{FH,SLIC,LSC,Manifold,SEAL,SSN,superpixel_fcn} do not explicitly enforce connectivity and there may exist some "orphaned" pixels that do not belong to the same connected components.
To correct this, SLIC~\cite{SLIC} assigns these pixels the label of the nearest cluster.
\cite{SSN,superpixel_fcn} also apply a component connection algorithm to merge superpixels that are smaller than a certain threshold with the surrounding ones.
These algorithms enforce connectivity using a post-processing step, whereas SNIC~\cite{SNIC} enforces connectivity explicitly from the start. SNIC uses a priority queue to choose the next pixel to be assigned, and the queue is populated with pixels which are 4 or 8-connected to a currently growing superpixel.
As far as we know, there is no method which utilizes learned features and enforces connectivity explicitly.

% \begin{figure}
%    \begin{center}
%    \includegraphics[width=0.4\linewidth]{./figure/initial.pdf}
%    \end{center}
%    \vspace{-7mm}
%       \caption{\textbf{Illustration of superpixel initialization.} 
%       % For each pixel in the green box, we consider the 2 grid cells in the red box for assignment. Left and right demonstrate neighboring superpixels in horizontal and vertical updating step respectively.
%       Left is a part of image pixels, green pixels represent sampled pixels. Right is a part of initial pixel-superpixel map, values of the map represent the ID of superpixels to which sampled pixels are assigned.}
%    \label{fig:initial}
%    \vspace{-20mm}
% \end{figure}   

% \begin{algorithm}[tb]
% \caption{SIN}
% \label{alg:algorithm}
% \textbf{Input}: Image $I$.\\
% \textbf{Output}: Superpixels $M$.
% \begin{algorithmic}[1] %[1] enables line numbers
% \STATE Let $t=0$.
% \WHILE{condition}
% \STATE Do some action.
% \IF {conditional}
% \STATE Perform task A.
% \ELSE
% \STATE Perform task B.
% \ENDIF
% \ENDWHILE
% \STATE \textbf{return} solution
% \end{algorithmic}
% \end{algorithm}

\section{Superpixel Segmentation Method}

\begin{figure}
   \begin{center}
   %  \vspace{-7mm}
   % \fbox{\rule{0pt}{2in} \rule{.9\linewidth}{0pt}}
   \includegraphics[width=1.0\linewidth]{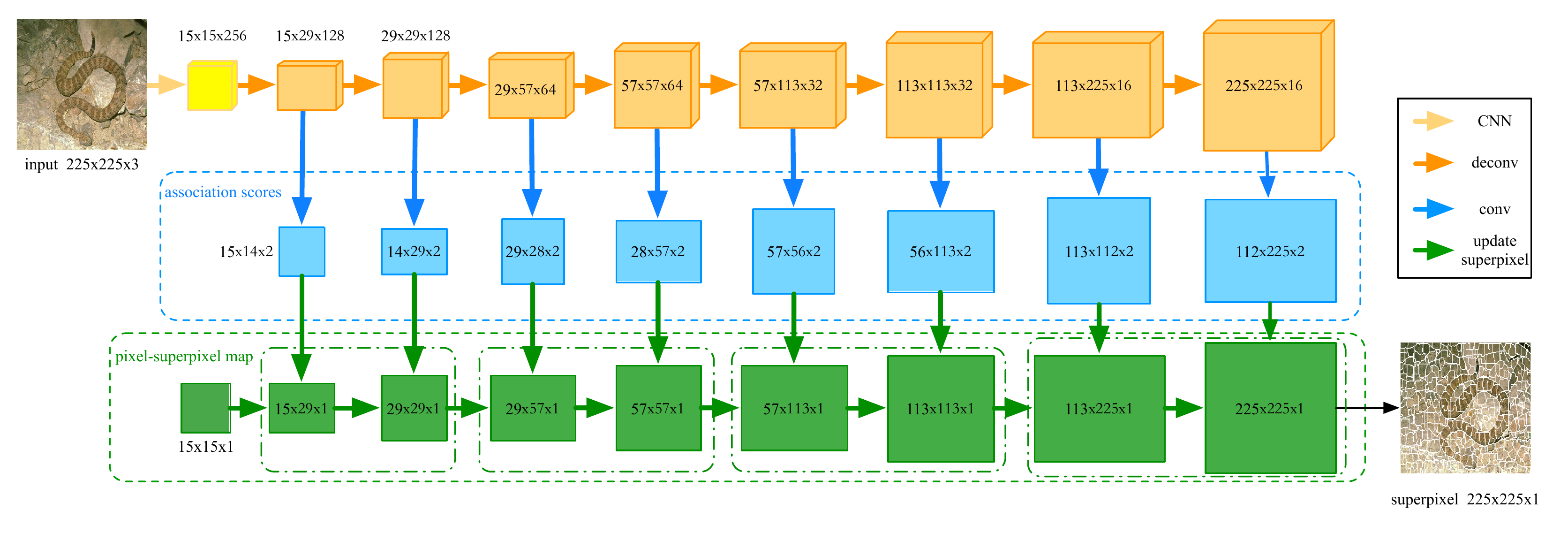}
   \vspace{-12mm}
   \end{center}
      \caption{\textbf{Illustration of our proposed method.} The SIN model takes the image as input, and predicts association scores for each updating step. In the training stage, association scores are utilized to compute loss. In the testing stage, new pixel-superpixel maps are obtained from current pixel-superpixel maps and association scores.}
   \label{fig:framework}
   \vspace{-5mm}
   \end{figure}

In this section, we introduce our superpixel segmentation method SIN.
The framework of our proposed method is illustrated in Figure~\ref{fig:framework}. 
We first present our idea of superpixel initialization and updating scheme.
After that, we introduce our network architecture and loss function design.
Finally, we will explain why our method can enforce spatial connectivity from the start.

%-------------------------------------------------------------------------
\subsection{Learn superpixels by Interpolation}
Our superpixels are obtained by initializing pixel-superpixel map and updating the map multiple times.
Similar to the commonly adopted strategy in \cite{SEEDS,SLIC,SNIC}, we generate the initial superpixels by sampling the image $I \in \mathbb{R}^{H\times W\times 3}$ with a regular step $S$. 
By assigning each pixel to a unique superpixel, we get the initial pixel-superpixel map $M_0 \in \mathbb{Z}^{h_0\times w_0}$. 
The values of $M_0$ denote ID of superpixels to which sampled pixels are assigned.

\begin{figure}
   \begin{center}
   % \fbox{\rule{0pt}{2in} \rule{.9\linewidth}{0pt}}
   \includegraphics[width=0.9\linewidth]{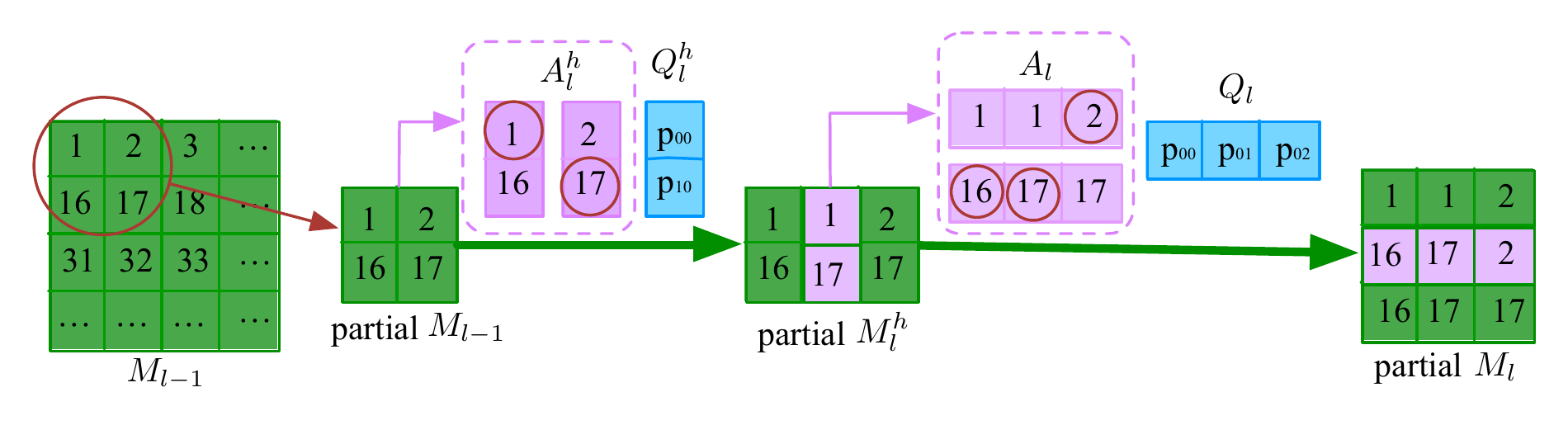}
   \vspace{-8mm}
   \end{center}
      \caption{\textbf{Illustration of expanding pixel-superpixel map.} Each expanding step consists of a horizontal interpolation followed by a vertical interpolation. The horizontal interpolation inserts values in each row and the vertical interpolation inserts values in each column. The inserted values are determined by association scores and neighboring superpixels.}
      % The SIN model takes the image as input, and predicts pixel-superpixel association scores at each update iteration. The association scores are then utilized to generate superpixels by assigning pixels to their horizontal or vertical neighboring superpixels.}
   \label{fig:update}
   \vspace{-3mm}
   \end{figure}

Superpixel segmentation is to find the final pixel-superpixel map $M \in \mathbb{Z}^{H\times W}$, which assigns all pixels to superpixels.
The problem of finding $M$ can be seemed as expanding $M_0$ to $M$. Inspired by resizing image, we use interpolation to expand the matrix.
The rule of interpolation is carefully designed to enforce spatial connectivity from the start and to be computed on GPU in parallel. 
As depicted in Figure\ref{fig:framework}, the process of expanding $M_0$ to $M$ can be divided into multiple similar steps and each step consists of a horizontal interpolation and a vertical interpolation.
As shown in Figure~\ref{fig:update}, when we expand pixel-superpixel map in horizontal/vertical dimension, we interpolate values among all neighboring elements in each row/column. The inserted values are the same as neighboring elements with certain probability. The probabilities(association scores) are computed by neural networks which we will introduce in Section~\ref{section:network}.

% We need to interpolate values into $M_0$ multiple times. Owing to $M_0$ has two dimensions, we expand superpixel map from horizontal and vertical dimensions at a interpolation. $M_l^h \in \mathbb{Z}^{h_{l-1}\times w_l}$ and $M_l\in \mathbb{Z}^{h_l\times w_l}$ denote superpixel maps after $l$-th horizontal and vertical interpolation.
% We update our superpixels by inserting superpixels into current superpixel map in horizontal dimension and vertical dimension in sequence. In general, we can use Equation~(\ref {general}) to describe the process of interpolation. $SP$ denotes the ID of superpixel inserted, $M$ denotes current superpixel map, $A$ denotes association scores which are computed by neural network described in Section~\ref{section:network}. $\mathcal{I}$ denotes the interpolation function.  We will demonstrate superpixel initialization and superpixel updating steps in detail in the following.
% \begin{equation}
%   SP = \mathcal{I} (M, A)
%    \label{general}
% \end{equation}
% After assigning these new inserted pixels to superpixel, we can get a new superpixel map. By iteratively updating the superpixel map, we can generate the final superpixels $M\in \mathbb{Z}^{H\times W}$.  

In detail, we use $P\in \mathbb{R}^{H\times W}$ to denote image pixels. 
% In this work, the index of a matrix starts from 0. 
$P(i,j)$ represents the image pixel at the intersection of $i$-th row and $j$-th column. $M(i,j)$ is the superpixel to which $P(i,j)$ is assigned.
In the initial step, we find partial connections between image pixels $P$ and superpixels.
$M_0(i,j)$ represents the superpixel to which $P(i*S,j*S)$ is assigned.
% We need to find the map between other pixels and the final superpixels $M$ by iteratively interpolating superpixels into current superpixel map.
\begin{equation}
    % \vspace{-2mm}
   h_0 = (H+S-1)/S, \quad
   w_0 = (W+S-1)/S.
   \label{h0}
%    \vspace{-2mm}
\end{equation}
To obtain $M$, we need to expand $M_0$ multiple times. At $l$-th expansion, we use $M_l^h\in \mathbb{Z}^{h_{l-1}\times w_l}$ and $M_l\in \mathbb{Z}^{h_l\times w_l}$ to denote pixel-superpixel maps after horizontal and vertical interpolation.
\begin{equation}
    % \vspace{-2mm}
   h_l = 2\ast h_{l-1} -1, \quad
   w_l = 2\ast w_{l-1} -1.
   \label{hl}
\end{equation}
% As shown in Figure~\ref{fig:update}, we interpolate superpixels in horizontal and vertical dimension step by step. To describe easily, we use one updating iteration to denote a horizontal interpolation and a vertical interpolation. We use $M_l^h\in \mathbb{Z}^{h_{l-1}\times w_l}$ and $M_l\in \mathbb{Z}^{h_l\times w_l}$ to denote superpixel maps after $l$-th horizontal and vertical dimensional interpolation, where $l$ is the index of current updating iteration.

Figure~\ref{fig:update} has shown a part of interpolation at $l$-th expansion.
At $l$-th horizontal/vertical interpolation step, the inserted values are confirmed by association scores $A_l^h\in \mathbb{R}^{h_{l-1}\times (w_{l-1}-1) \times 2}$/$A_l\in \mathbb{R}^{(h_{l-1}-1)\times w_l \times 2}$ and neighboring superpixels $Q_l^h \in \mathbb{Z}^{h_{l-1}\times (w_{l-1}-1) \times 2}$/$Q_l\in \mathbb{Z}^{(h_{l-1}-1)\times w_l \times 2}$.
$A_l^h(i,j,k)$ and $A_l(i,j,k)$ denote the probability of $i$-th row, $j$-th column inserted value is the same with its $k$-th neighbor.
All association scores are obtained from multi-layer outputs of the neural network described in Section~\ref{section:network}.
$Q_l^h(i,j,k)$ and $Q_l(i,j,k)$ denote the $k$-th neighbor's value of $i$-th row, $j$-th column inserted element.
Neighboring superpixels are obtained from current pixel-superpixel map. We interpolate new elements among existing neighboring elements at each row/column, so a pair of existing neighboring elements' values are neighboring superpixel IDs of the corresponding inserted element. According to association scores and neighboring superpixels, inserted values can only be same with one of their neighboring elements with certain probability.

\subsection{Network Architecture and Loss Function} \label{section:network}

We use a convolutional neural network similar to \cite{superpixel_fcn} to extract image feature $F_0\in \mathbb{R}^{h_0\times w_0\times c_0}$. We stack module $deconv\_h$ and $deconv\_v$ multiple times to extract multi-layer features $F_l^h \in \mathbb{R}^{h_{l-1}\times w_l\times c_l}$ and $F_l\in \mathbb{R}^{h_l\times w_l\times c_l}$,
% \begin{equation}
%     c_l = c_{l-1}/2
%     \label{cl}
%  \end{equation}
% and initialize superpixel map $M_0 \in \mathbb{Z}^{h_0\times w_0}$.
% Different with \cite{superpixel_fcn}, we do not utilize feature from previous layers to predict scores.
% As Figure~\ref{fig:framework} shown, we get $Q_l^h$ and $Q_l$ through $deconv\_h$ and $deconv\_v$ followed by $conv$.
where $c_l$ denotes feature channels.
$deconv\_h$ and $deconv\_v$ are transposed convolutional neural networks, in which stride are $(1, 2)$ and $(2, 1)$ respectively. Specially, $deconv\_h$ will reduce feature channels by half.
$conv$ is a convolutional neural network, which transforms the multi-layer features to 2-dimensional association scores.

Our model is trained with ground truth segmentation labels $T\in \mathbb{Z}^{H\times W}$ from BSDS500.
Every interpolation is to find partial connections between pixels and superpixels. To get loss of all connections, we need to compute partial loss at every interpolation.
We define $s_l=S/2^l$ to simplify descriptions.
The inserted values at $l$-th step in horizontal/vertical dimension are ID of superpixels to which pixels $U_l^h$ and $U_l$ are assigned. $U_l^h$ denotes the subtraction of pixels sampled by stride $(s_{l-1},s_{l})$ and $(s_{l-1},s_{l-1})$. $U_l$ denotes the subtraction of pixels sampled by stride $(s_{l},s_{l})$ and $(s_{l-1},s_{l})$. Partial ground truth connections $T_l^h$ and $T_l$ are segmentation labels of pixels $U_l^h$ and $U_l$.
To speed up training process, we do not generate pixel-superpixel maps to compute loss.
Instead, we utilize association scores to compute loss directly.
Association scores $A_l^h$ and $A_l$ denote the probabilities of pixels assigned to neighboring superpixels.
Inspired by tasks of classification, the ground truth labels $G_l^h$ and $G_l$ are defined as the indexes of neighboring superpixels to which pixels should be assigned. $G_l^h$ and $G_l$ can be inferred from $T_l^h$ and $T_l$. Owing to each inserted element has two neighbors, the ground truth labels are 0 or 1.
If the neighboring superpixels ID of an inserted element are same, we will ignore it when computing loss. We define $\mathbb{I}_l^h$ and $\mathbb{I}_l$ to represent whether to consider the elements when computing loss. Loss of each interpolation at $l$-th step can be computed by:
\begin{equation}
   L_l^h = \mathcal{C}_{\mathbb{I}_l^h}(G_l^h, A_l^h), \quad
   L_l^v = \mathcal{C}_{\mathbb{I}_l}(G_l, A_l)
\end{equation}
where $L_l^h$ and $L_l$ denote the loss of horizontal and vertical at $l$-th step. $\mathcal{C}_{\mathbb{I}_l^h}$ and $\mathcal{C}_{\mathbb{I}_l}$ denote cross entropy loss functions, which only consider partial elements according to the values of $\mathbb{I}_l^h$ and $\mathbb{I}_l$.

Total loss $\mathcal{L}$ can be computed by:
\begin{equation}
   \mathcal{L} = -\sum_{l}\left( w_l^h L_l^h+ w_l^v L_l^v \right)
\end{equation}
where $w_l^h$ and $w_l^v$ denote weights of horizontal and vertical interpolation loss at $l$-th step.

\subsection{Illustration of Spatial Connectivity}
\label{section:connectivity}
Thanks to removing the post-processing step, our method can be integrated into subsequent tasks in an end-to-end way.
The key to enforcing spatial connectivity from the start is the rule of interpolation.
An expanding step consists of a horizontal interpolation and a vertical interpolation. The design ensures spatial connectivity of pixel-superpixel maps will not be destroyed by interpolations. Owing to initial pixel-superpixel map has spatial connectivity and interpolations preserve the property, the final pixel-superpixel map $M$ remains spatial connectivity.
$M$ assigns all pixels to superpixels, so $M$ has spatial connectivity equals our SIN superpixels have spatial connectivity. In the following, we will first explain why the spatial connectivity of $M$ and superpixels are equivalent. Afterwards, we illustrate how interpolations preserve spatial connectivity of pixel-superpixel maps.

The fact that a superpixel has spatial connectivity means the set of all pixels in the superpixel is a connected set. We use $X_i$ to denote a set where elements have same value $i$ in $M$ and $X=\{X_1, X_2, \dots, X_n\}$ to denote all such sets. If all elements in $X$ are connected sets, $M$ has spatial connectivity. Spatial information of elements in $X_i$ equals spatial information of pixels assigned to superpixel $i$, so $X_i$ is a connected set represents superpixel $i$ has spatial connectivity. Evidently, $M$ has spatial connectivity equals all superpixels have spatial connectivity. All sets in $M_0$ only has one element, so $M_0$ has spatial connectivity definitely. If interpolations can preserve spatial connectivity, we can infer that $M$ has spatial connectivity.

Our scheme of interpolation is to insert elements among existing neighboring elements at each row/column. When we insert a element between a pair of neighbors, only sets including these three elements will be taken into consideration. If existing neighboring elements are in a same set, the inserted element will be added to the set and the set is still connected. If existing neighboring elements belong to different sets, the inserted element will be added to one of the sets, and the other will not change. The added set is still connected and spatial connectivity of the other will not be affected. We want to address that it is the design of interpolation preserves spatial connectivity. If we interpolate once at an expanding step and the inserted value is same with its 8-neighborhood, spatial connectivity of pixel-superpixel map will be destroyed.
Above all, our method can enforce spatial connectivity explicitly through the delicate design of interpolation.

% Final pixel-superpixel map $M$ assigns all pixels to superpixels, so the set of elements having same value 
% in $M$ has same spatial information with the corresponding superpixel. We use $X_i$ to denote a set where elements have same value in $M$ and $X=\{X_1, X_2, \dots, X_n\}$ to denote all such sets in $M$. If all sets in $X$ have spatial connectivity, superpixels have spatial connectivity as well. The set $X_i$ has spatial connectivity equals to $X_i$ is a connected set. $M$ is obtained by expanding $M_0$ as same way for multiple times. According to recursion, if $M_0$ only has connected sets and if $M_{l-1}$ only has connected sets then $M_l$ only has connected sets, we can inference that $M$ only has connected sets.

% Any $X_i^0 \in X^0$ has one element, so it is a connected set definitely. When we expand $M_{l-1}$ to $M_l$, a horizontal and a vertical interpolation are operated. At a horizontal/vertical interpolation, the inserted element can only affect the connectivity of its two neighboring elements and itself. If these two elements are in a same set, the inserted will become a part of the set. If these two elements are in two different sets, the inserted element will become a part of one of the sets, and the other set won't be affected. So all sets of $M_l$ are connected sets. We can infer all sets in $M$ are connected sets. So our method can enforce spatial connectivity of superpixels from the start.

%------------------------------------------------------------------------
\vspace{-3mm}
\section{Experiments}
\begin{figure*}[t]
   \centering
%    \vspace{-7mm}
   % \begin{tabular}{cccccc}
   % \hspace{-0.5mm}\includegraphics[height =0.77in]{Input} & GT segments & SLIC & SEAL & SSN & Ours
   % \end{tabular}
   \begin{tabular}{ccccccc}
   %\hspace{-0.5mm}\includegraphics[height =0.41in]{figures/scheme1.pdf} &
   \hspace{-0.5mm}Input & \hspace{-0.5mm}GT & \hspace{-0.5mm}SNIC & \hspace{-0.5mm}SEAL & \hspace{-0.5mm}SSN & \hspace{-0.5mm}FCN & \hspace{-0.5mm}Ours\\
   \hspace{-0.5mm}\includegraphics[height =0.444in]{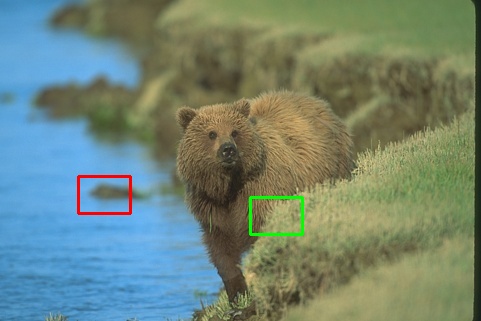} &
   \hspace{-0.5mm}\includegraphics[height= 0.444in]{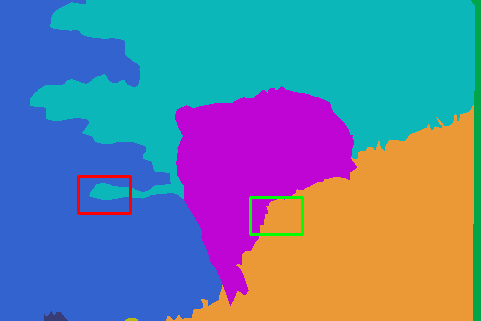} &
   \hspace{-0.5mm}\includegraphics[height= 0.444in]{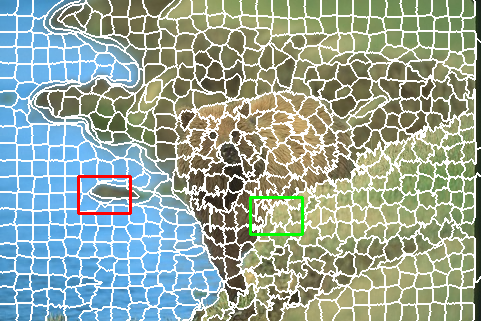} &
   \hspace{-0.5mm}\includegraphics[height =0.444in]{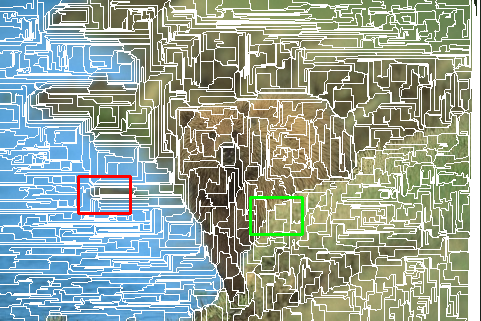} &
   \hspace{-0.5mm}\includegraphics[height =0.444in]{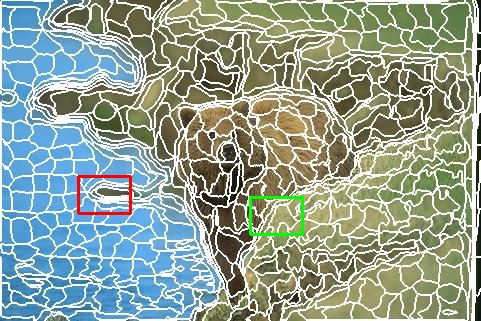} &
   \hspace{-0.5mm}\includegraphics[height =0.444in]{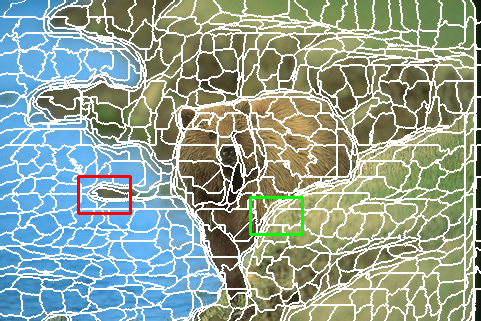} &
   \hspace{-0.5mm}\includegraphics[height =0.444in]{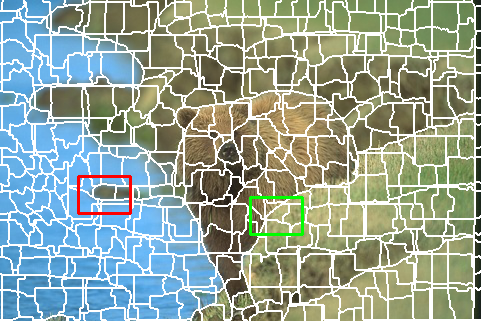} 
   \end{tabular}
   \begin{tabular}{cccccccccccccc}
   %\hspace{-0.5mm}\includegraphics[height =0.41in]{figures/scheme1.pdf} &
   \hspace{-0.5mm}\includegraphics[height =0.23in]{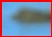} &
   \hspace{-0.5mm}\includegraphics[height =0.23in]{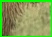} &
   \hspace{-0.5mm}\includegraphics[height= 0.23in]{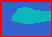} &
   \hspace{-0.5mm}\includegraphics[height= 0.23in]{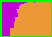} &
   \hspace{-0.5mm}\includegraphics[height =0.23in]{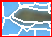} &
   \hspace{-0.5mm}\includegraphics[height =0.23in]{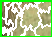} &
   \hspace{-0.5mm}\includegraphics[height =0.23in]{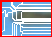} &
   \hspace{-0.5mm}\includegraphics[height =0.23in]{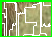} &
   \hspace{-0.5mm}\includegraphics[height =0.23in]{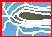} &
   \hspace{-0.5mm}\includegraphics[height =0.23in]{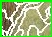} &
   \hspace{-0.5mm}\includegraphics[height =0.23in]{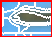} &
   \hspace{-0.5mm}\includegraphics[height =0.23in]{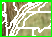} &
   \hspace{-0.5mm}\includegraphics[height =0.23in]{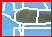} &
   \hspace{-0.5mm}\includegraphics[height =0.23in]{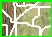}
   \end{tabular}
   \begin{tabular}{ccccccc}
   %\hspace{-0.5mm}\includegraphics[height =0.41in]{figures/scheme1.pdf} &
   % \hspace{-0.5mm}Input & GT segments & SLIC & SEAL & SSN & Ours\\
   \hspace{-0.5mm}\includegraphics[height =0.49in]{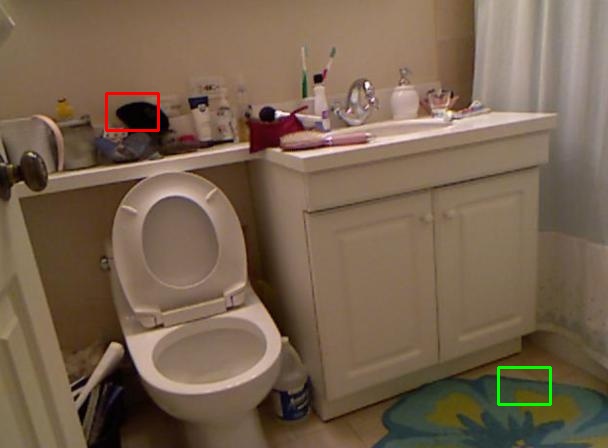}&
   \hspace{-0.5mm}\includegraphics[height= 0.49in]{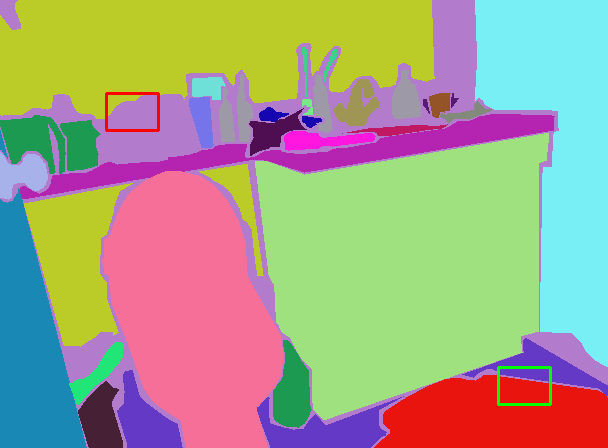} &
   \hspace{-0.5mm}\includegraphics[height =0.49in]{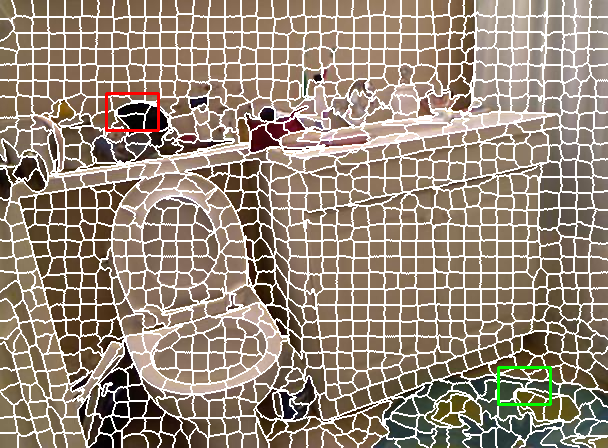} &
   \hspace{-0.5mm}\includegraphics[height =0.49in]{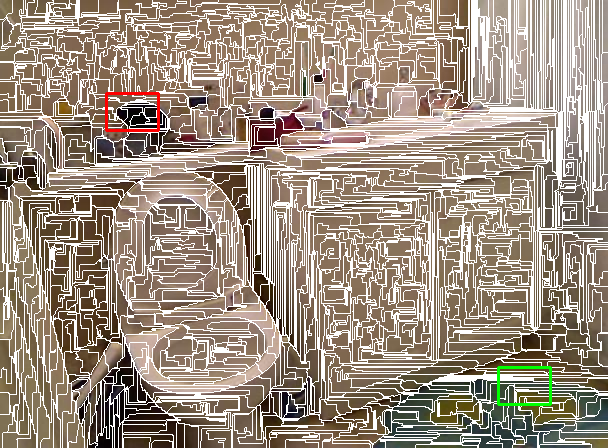} &
   \hspace{-0.5mm}\includegraphics[height =0.49in]{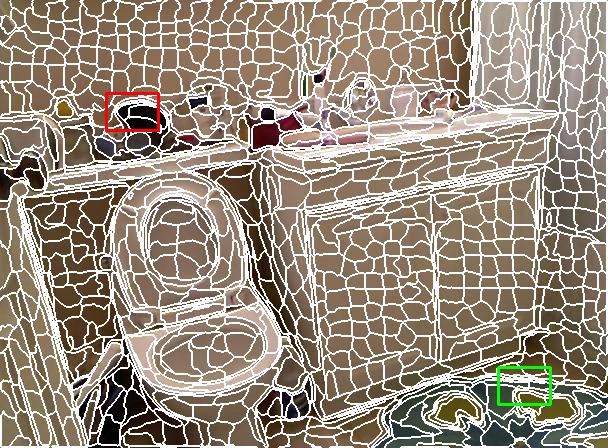} &
   \hspace{-0.5mm}\includegraphics[height =0.49in]{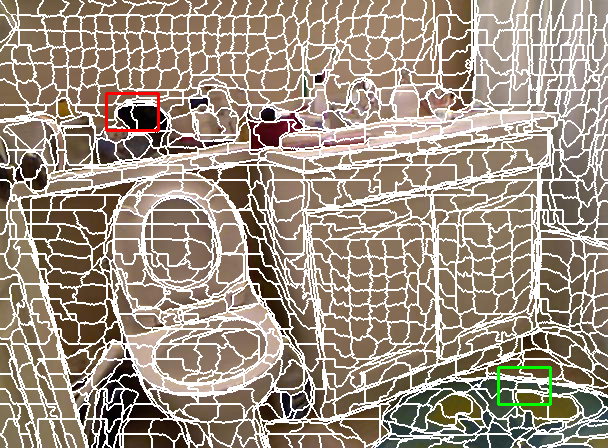} &
   \hspace{-0.5mm}\includegraphics[height =0.49in]{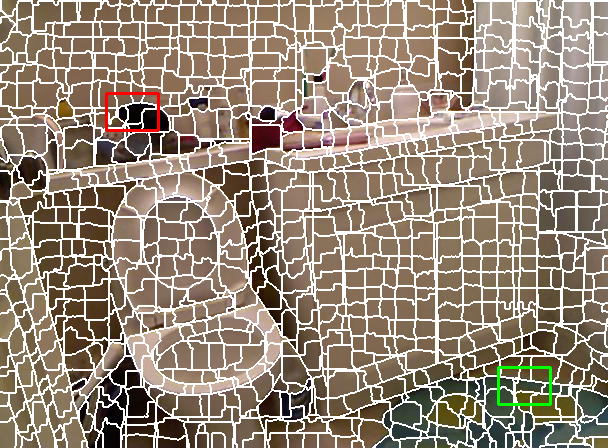}
   \end{tabular}
   \begin{tabular}{cccccccccccccc}
      %\hspace{-0.5mm}\includegraphics[height =0.41in]{figures/scheme1.pdf} &
      \hspace{-0.5mm}\includegraphics[height =0.23in]{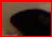} &
      \hspace{-0.5mm}\includegraphics[height =0.23in]{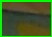} &
      \hspace{-0.5mm}\includegraphics[height= 0.23in]{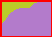} &
      \hspace{-0.5mm}\includegraphics[height= 0.23in]{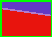} &
      \hspace{-0.5mm}\includegraphics[height =0.23in]{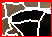} &
      \hspace{-0.5mm}\includegraphics[height =0.23in]{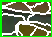} &
      \hspace{-0.5mm}\includegraphics[height =0.23in]{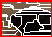} &
      \hspace{-0.5mm}\includegraphics[height =0.23in]{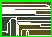} &
      \hspace{-0.5mm}\includegraphics[height =0.23in]{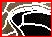} &
      \hspace{-0.5mm}\includegraphics[height =0.23in]{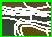} &
      \hspace{-0.5mm}\includegraphics[height =0.23in]{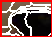} &
      \hspace{-0.5mm}\includegraphics[height =0.23in]{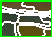} &
      \hspace{-0.5mm}\includegraphics[height =0.23in]{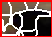} &
      \hspace{-0.5mm}\includegraphics[height =0.23in]{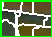}
      \end{tabular}
   \vspace{-6mm}
   \caption{\textbf{Visual results.} Compared to SEAL, SSN and FCN, our method is competitive or better in terms of object boundary adherence while generating more compact superpixels. Top rows: BSDS500. Bottom rows: NYUv2. }\vspace{-5mm}
   \label{fig:spixel_viz}
   \end{figure*}

\vspace{-3mm}
To be integrated into subsequent tasks in an end-to-end way without impeding their instantaneity, we analyze the runtime of deep learning-based models. To demonstrate the effectiveness of SIN in superpixel segmentation, we train and test our model on the standard benchmark BSDS500~\cite{BSD}.
We also report its performance without fine-tuning on the benchmark NYUv2~\cite{NYU} to evaluate the generalizability of our model.
We use protocols and codes provided by \cite{evaluation} to evaluate all methods on two benchmarks. SNIC~\cite{SNIC}, SEAL~\cite{SEAL}, SSN~\cite{SSN} and FCN~\cite{superpixel_fcn} are tested with the original implementations from the authors. SLIC~\cite{SLIC} and ERS~\cite{ERS} are tested with the codes provided in \cite{evaluation}. For SLIC and ERS, we use the best parameters reported in \cite{evaluation}, and for the rest, we use the default parameters recommended in the original papers. Figure~\ref{fig:spixel_viz} shows the visual results of some state-of-the-art methods and ours.

\vspace{-3mm}
\subsection{Comparison with the state-of-the-arts}

\paragraph{Implementation details.} We implement our model with PyTorch and use Adam with $\beta_1=0.9$ and $\beta_2=0.999$ to optimize it. For training, we randomly crop the images to size $225\times 225$ as input and perform horizontal/vertical flipping for data augmentation. The initial learning rate is set to $5\times 10^{-5}$ and is reduced by half after 200k iterations. It takes us about 3 hours to train the model for 300k iterations on 1 NVIDIA RTX 2080Ti GPU device. 

We set the regular step $S$ as 16 and we can get $15\times 15$(225) superpixels through 4 expanding steps when training. We set $w^h$ and $w^v$ as $[20, 10, 5, 2.5]$ and $[8, 4, 2, 1]$ respectively. To generate the varying number of superpixels when testing, we simply resize the input image to the appropriate size. For example, if we want to generate $30\times 20$ superpixels, we can resize the image to $(30*16-15)\times (20*16-15)$ 
\emph{i.e.} $465\times305$.

% ===== time
\begin{wrapfigure}{r}{4cm}
    \centering
    \vspace{-5mm}
    \includegraphics[height =1.2in]{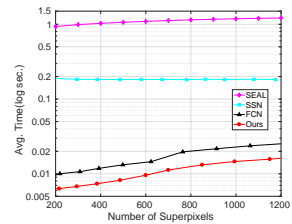}
    \vspace{-7mm} 
    \caption{\textbf{Runtime analysis.} Average runtime of different DL methods \emph{w.r.t} number of superpixels. Note that $y$-axis is plotted in the logarithmic scale.}
    % \vspace{-4mm}
    \label{fig:runtime}
    \end{wrapfigure}
    % === 

\paragraph{Runtime Analysis.} We compare the runtime difference between deep learning-based methods. Figure~\ref{fig:runtime} reports the average runtime \emph{w.r.t} the number of generated superpixels on a NVIDIA RTX 2080Ti GPU device. Our method runs about 1.5 to 2 times faster than FCN, 12 to 33 times faster than SSN, and more than 70 times faster than SEAL. Note that existing deep learning-based methods need a post-processing step which takes 2.5ms to 8ms \cite{gslic} and runtime in Figure~\ref{fig:runtime} does not include the time. The reason of our method has the fastest speed is that we use a novel interpolation method to generate superpixels. What's more, our method saves plenty of training time compared to FCN due to the simple and effective loss function. For training, we spend about 3 hours on a single GPU, while FCN spends about 20 hours. 

\paragraph{Evaluation metrics.} To demonstrate the effectiveness of SIN, we use the achievable segmentation accuracy (ASA), boundary recall and precision (BR-BP), and compactness (CO) to evaluate the superpixels. ASA evaluates superpixels by measuring the total effective segmentation area of a superpixel representation concerning the ground truth segmentation map. BR and BP measure the boundary adherence of superpixels given the ground truth boundary, whereas CO assesses the compactness of superpixels. The higher these scores are, the better the superpixel segmentation result is. As in \cite{evaluation}, for BR and BP evaluation, the boundary tolerance is 0.0025 times the image diagonal rounded to the closest integer.

\paragraph{Results on BSDS500.} BSDS500 contains 200 training, 100 validation, and 200 test images. Each image in this dataset is provided with multiple ground truth annotations. For training, we follow \cite{SSN,SEAL,superpixel_fcn} and treat each annotation as an individual sample. With this dataset, we have 1633 training/validation samples and 1063 testing samples. We train our model using both the training and validation samples.

\begin{figure*}[t]
   \centering
   \begin{tabular}{ccc}
   %\hspace{-0.5mm}\includegraphics[height =0.41in]{figures/scheme1.pdf} &
   \includegraphics[height =1.2in]{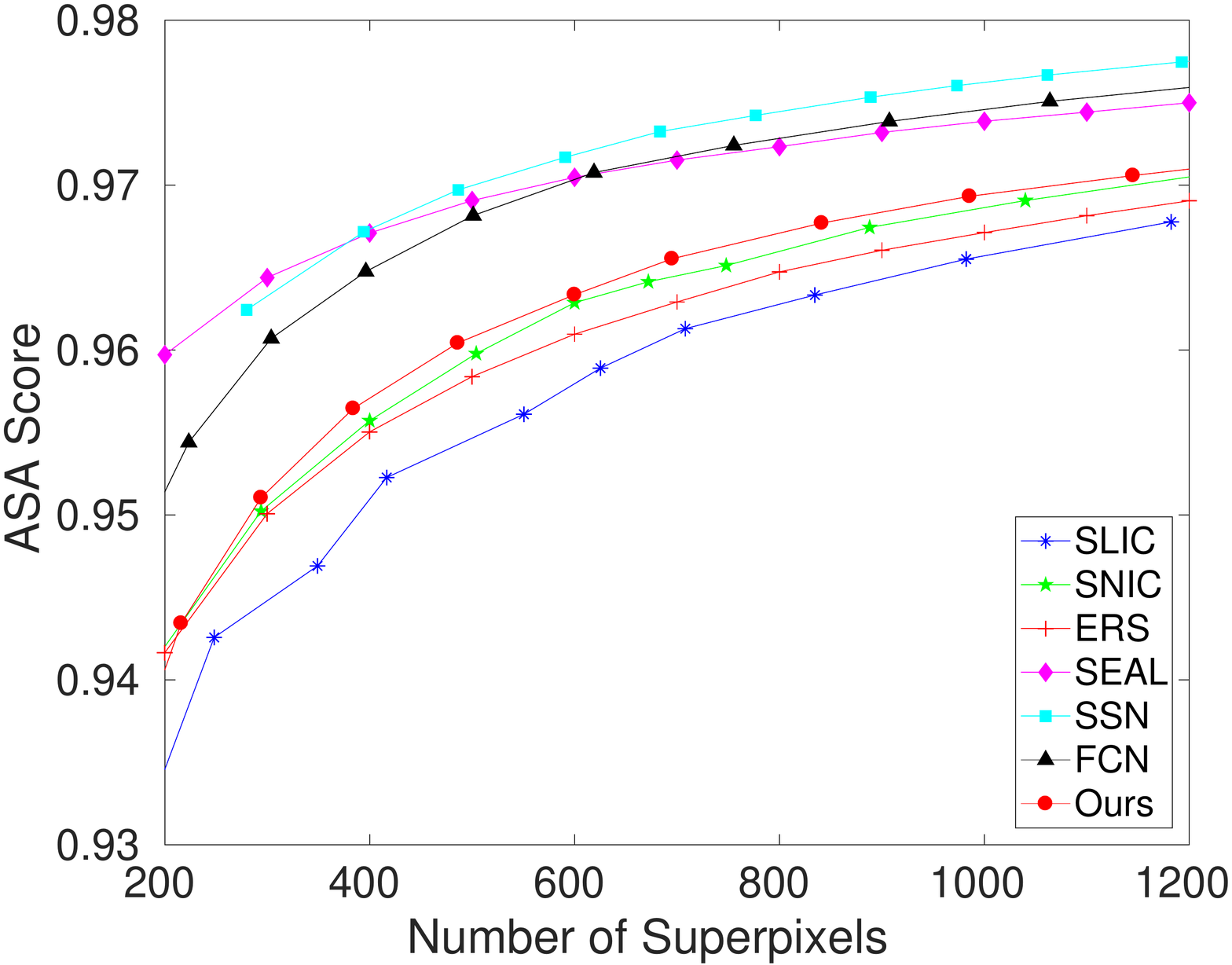} &
   \includegraphics[height =1.2in]{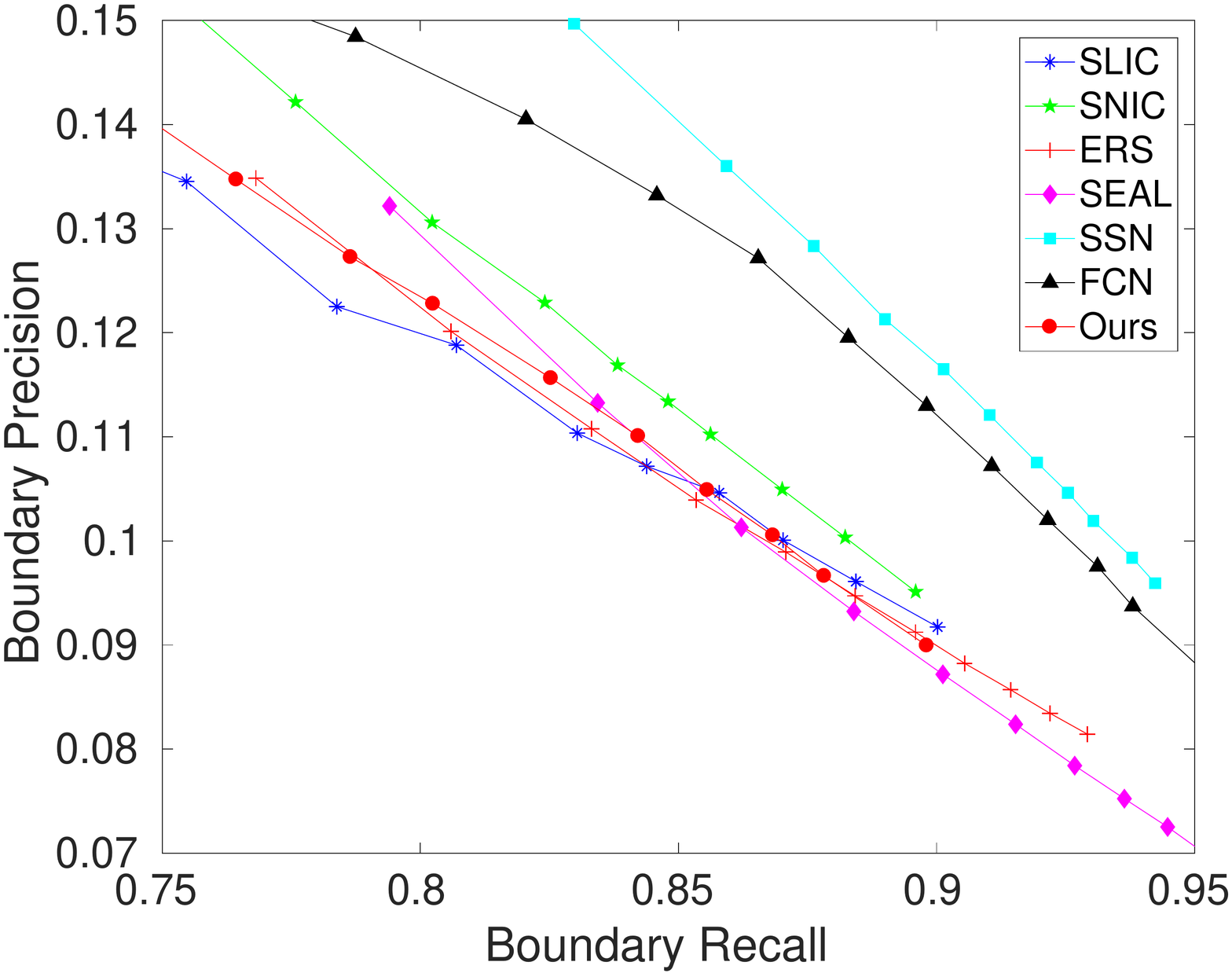} &
   \includegraphics[height =1.2in]{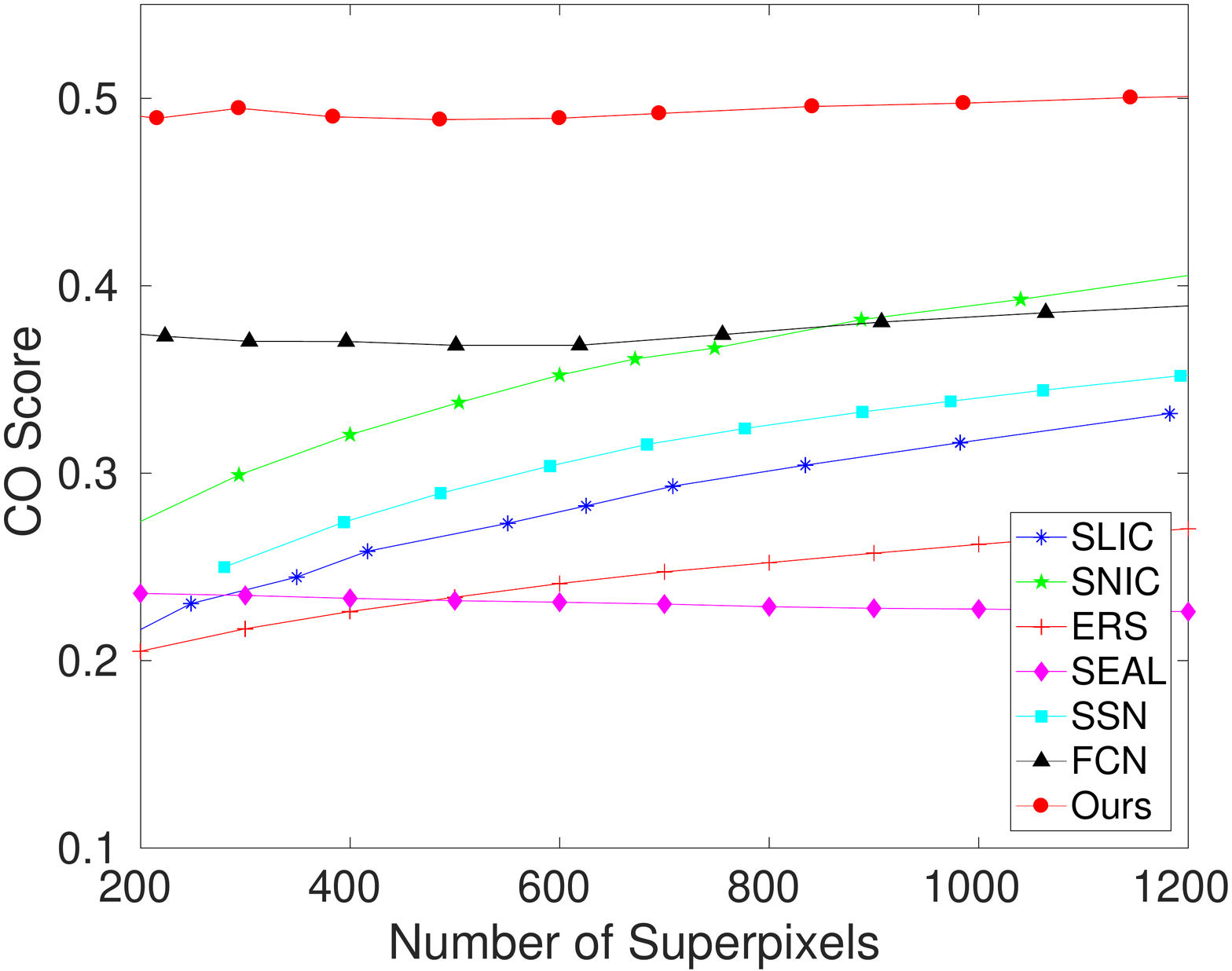}
   \end{tabular}
   \vspace{-3mm}
   \caption{\textbf{Results on BSDS500.} From left to right: ASA, BR-BP, and CO.}
   % \vspace{-3mm}
   \label{fig:BSDS_res}
   \end{figure*}

Figure~\ref{fig:BSDS_res} reports the performance of all methods on BSDS500 test set. Our method outperforms all traditional methods on all evaluation metrics, except SNIC in terms of BR-BP. Comparing to the other deep learning-based methods, our method achieves competitive results in terms of ASA and BR-BP, and signiﬁcantly higher scores in terms of CO. With high CO, our method can better capture spatially coherent information and avoids paying too much attention to image details and noises. As shown in Figure~\ref{fig:spixel_viz}, when handling fuzzy boundaries, our method can generate smoother superpixels.   

\paragraph{Results on NYUv2.} NYUv2 is an RGB-D dataset containing 1499 images with object instance labels,  which is originally proposed for indoor scene understanding tasks. \cite{evaluation} removes the unlabelled regions near the image boundary and develops a benchmark on a subset of 400 test images with size $608 \times 448$ for superpixel evaluation. We directly apply the models of SEAL, SSN, FCN, and our method trained on BSDS500 to this dataset without any ﬁne-tuning.

\begin{figure*}[t]
   \centering
   \begin{tabular}{ccc}
   %\hspace{-0.5mm}\includegraphics[height =0.41in]{figures/scheme1.pdf} &
   \includegraphics[height =1.2in]{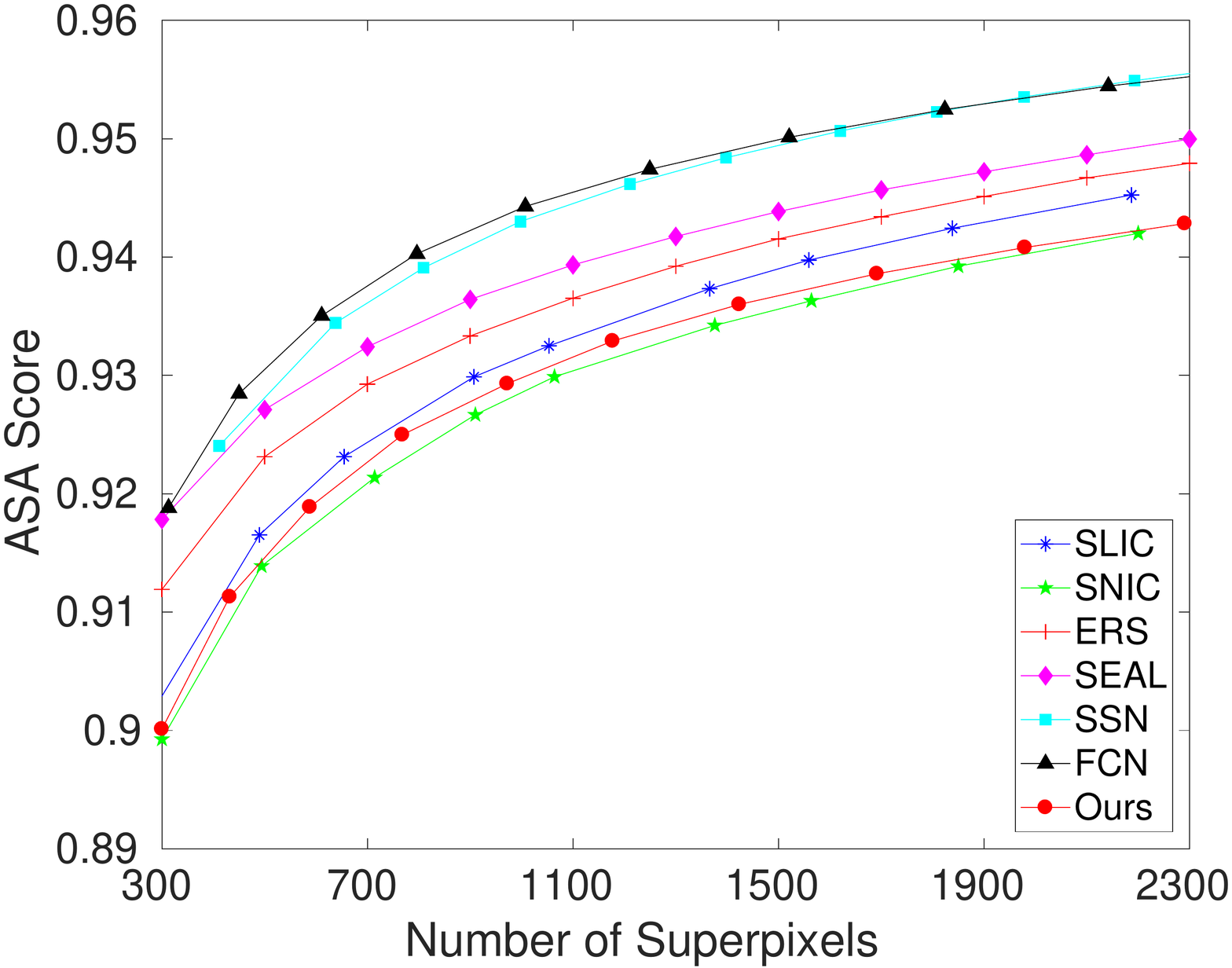} &
   \includegraphics[height =1.2in]{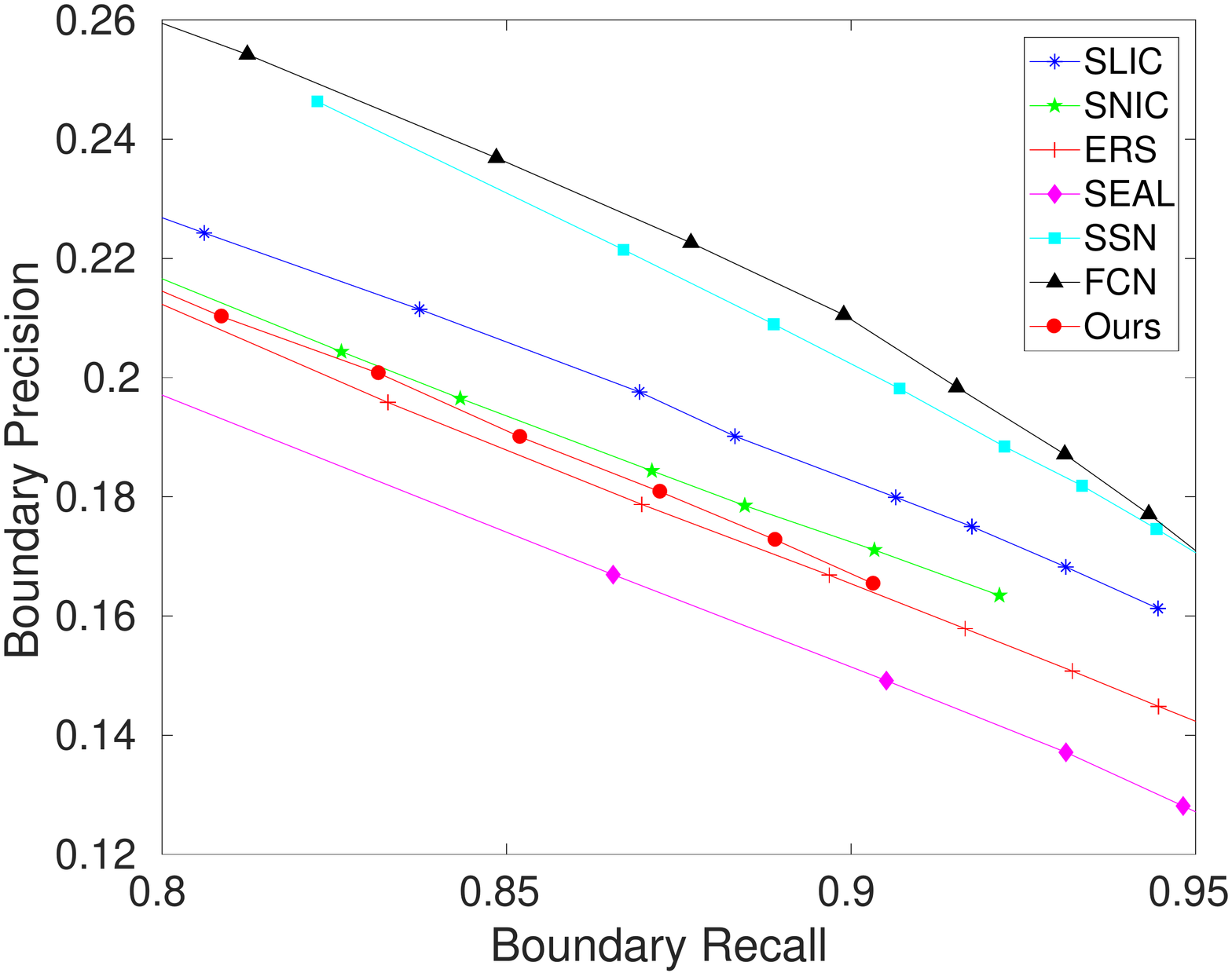} &
   \includegraphics[height =1.2in]{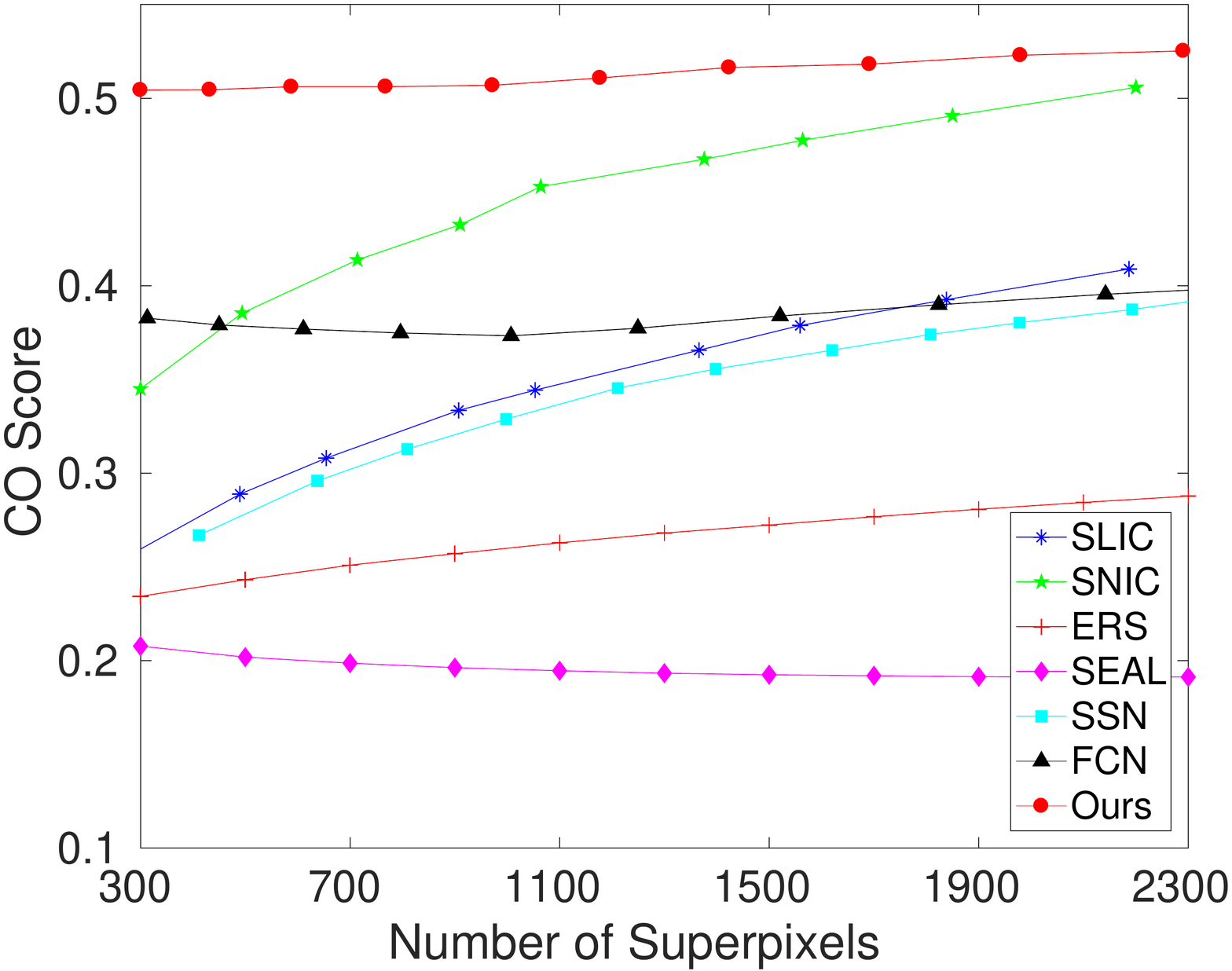}
   \end{tabular}
   \vspace{-3mm}
   \caption{\textbf{Results on NYUv2.} From left to right: ASA, BR-BP, and CO.}\vspace{-4mm}
   \label{fig:NYU_res}
   \end{figure*}

Figure~\ref{fig:NYU_res} shows the performance of all methods on NYUv2. In general, these deep learning-based algorithms achieve competitive or better performance against the traditional algorithms, which demonstrate that they can extract high-quality superpixels on other datasets. Also, our method outperforms all other methods in terms of CO. As the visual results shown in Figure~\ref{fig:spixel_viz}, our method handles the fuzzy boundary better than other deep learning-based methods.

\paragraph{Illustration of high CO score.} The experimental results on BSDS500 and NYUv2 show that our method has lower ASA and BR-BP scores, while a higher CO score. We will illustrate the reason in the following.

% \begin{wrapfigure}{r}{4cm}
\begin{figure}
\centering
\vspace{-5mm}
\includegraphics[height =1.5in]{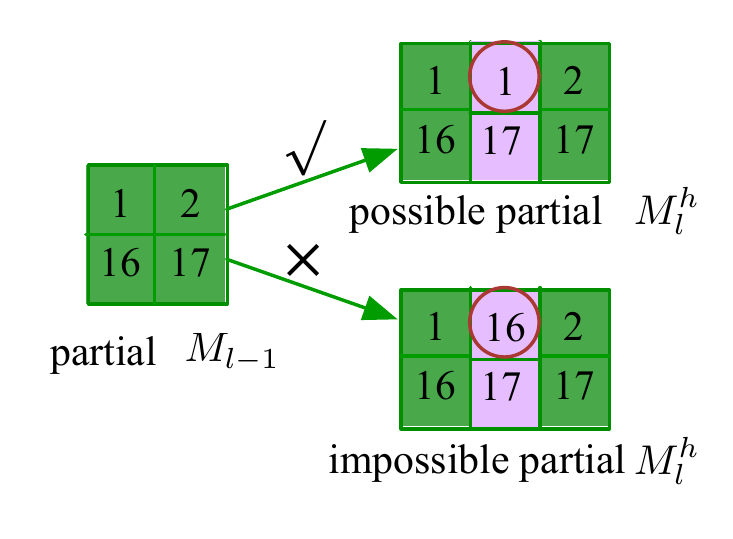}
\vspace{-5mm} 
\caption{\textbf{Illustration of high CO score.} According to the rule of interpolation, the above is a possible new pixel-superpixel map and the below is an impossible one.}
% \vspace{-7mm}
\label{fig:high_co}
% \end{wrapfigure}
\end{figure}
% === 

To enforce spatial connectivity from the first, we expand pixel-superpixel map in horizontal and vertical dimensions.
The horizontal/vertical interpolation constrains the inserted value can only be same with its horizontal/vertical neighbors.
As Figure~\ref{fig:high_co} shown, the black circled value can be 1/2 and cannot be 16/17. However, if the ground truth of the value is 16, our method cannot interpolate the same value. That is the reason our ASA and BR-BP scores are lower than other deep learning-based methods.
Meanwhile, the constraint results in pixels in a superpixel are 4-neighborhood connected which is more compact than 8-neighborhood connected.
Owing to the high CO score, our method generates smoother superpixels on the fuzzy boundaries as Figure~\ref{fig:spixel_viz} shown.
The importance of compactness has been demonstrated in \cite{superpixel_fcn}. To extract more useful features in downstream tasks, it is important to capture spatial coherence in the local region in our superpixel method.
In our view, it is worthy  to enforce spatial connectivity from the start and get a higher CO score while sacrificing slight ASA and BR-BP scores.

\subsection{Ablation study}

\begin{figure}
      \centering
      \vspace{-5mm}
      \includegraphics[height =1.6in]{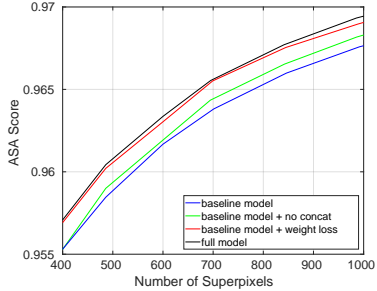}
      \vspace{-3mm} 
      \caption{\textbf{Ablation study.} We show the effectiveness of each design choice in the SIN model in improving accuracy.}
      \vspace{-5mm}
      \label{fig:ablation}
     %  \end{wrapfigure}
  \end{figure}

We present an ablation study where we evaluate different design choices of the image feature extraction and loss sum. Unlike \cite{superpixel_fcn}, we do not take image features from previous layers into account to predict association scores. Our total loss is the sum of horizontal and vertical loss at each step, so we can compute average or weighted sum. In our final model, we choose weighted sum to compute total loss.
% Unlike \cite{superpixel_fcn}, we do not take image features from previous layers into account to predict association scores. When computing the total loss of a whole superpixel extraction process, we assign weights to loss from each iteration and calculate the weighted sum. 
For comparison, we include a baseline model which uses the previous features and current features(concat) to predict scores and simply sums the loss values averagely. We evaluate each of these design options of the network. 
Figure~\ref{fig:ablation} shows that each of the 2 alternatives in our model performs better. 

% ===== ablation
% \begin{wrapfigure}{r}{4cm}
   %    \vspace{-3mm}
      % === 

\section{Application}
% \vspace{-3mm}

In this section, we evaluate whether our SIN superpixels can improve the performance of downstream vision tasks which utilize superpixels.
For this study, we choose existing semantic segmentation and salient object detection algorithms and substitute the original superpixels with our superpixels. For the following two tasks, our superpixels are generated by the network fine-tuned on PascaVOC 2012 training and validation datasets.

\paragraph{Semantic segmentation.}For semantic segmentation, CNN models \cite{CRF,fully4seg} achieve the state-of-the-art performance. However, most CNN architectures generate lower resolution outputs and then upsample them using post-processing techniques. To alleviate the need for post-processing CRF techniques, \cite{BI} propose the Bilateral Inception (BI) networks to utilize SLIC superpixels for long-range and edge-aware propagation across CNN units. 
We use SNIC and our superpixels to substitute SLIC superpixels and set the number of superpixels as 600. We evaluate the generated semantic segmentation on the PascalVOC 2012 test set~\cite{VOC}. Table~\ref{tab:voc} shows the standard Intersection over Union (IoU) scores. The results indicate that we can obtain signiﬁcant IoU improvements when using SIN superpixels.

\paragraph{Salient object detection.} Superpixels are widely used in salient object detection algorithms.
We experiment with Saliency Optimization(SO)~\cite{SO} and report standard Mean Absolute Error (MAE) scores on the ECSSD dataset~\cite{ecssd}.
To demonstrate the potential of our SIN Superpixels, we replace SLIC superpixels used in SO with ours, SNIC, and ERS superpixels and set the number of superpixels as 200 and 400.  Experimental results in Table~\ref{tab:ecssd} show that the use of our 200/400 superpixels consistently improves the performance of SO.

The above results on semantic segmentation and salient object detection demonstrate the effectiveness of integrating our superpixels into downstream vision tasks.

% \begin{table}[t]
%    \centering
%    \caption{\textbf{Superpixels for semantic segmentation.}We compute semantic segmentation using the BI network~\cite{BI} with different types of superpixels and compare the IoU scores on the PascalVOC 2012 test set.}
%    \vspace{3mm}
%    \label{tab:voc}
%    \begin{tabular}{lcc}
%    \hline
%    Method  & \# of Superpixels &IoU  \\
%    \hline 
%    DeepLab~\cite{CRF}    &   - &  68.9  \\
%    + CRF~\cite{CRF}   &  - & 72.7 \\
%    + BI(SLIC)~\cite{BI}  &  600 & 73.5 \\
%    + BI(ERS)  & 600 & 74.0 \\
%    + BI(Ours) & 600 & \textbf{74.4} \\
%    \hline
%    \end{tabular}
%    \vspace{-3mm}
%    \end{table}

\begin{table}
   \centering
% \vspace{-4mm}
\caption{\textbf{Superpixels for semantic segmentation.} We compute semantic segmentation using the BI network with different types of superpixels and compare the IoU scores on the PascalVOC 2012 test set.}
% \vspace{-2mm}
\label{tab:voc}
\begin{tabular}{|c|c|c|c|c|c|}
% \toprule
\hline
\hspace{0.1mm} Method  \hspace{0.1mm}& \hspace{0.1mm} DeepLab~\cite{CRF} \hspace{0.1mm} & \hspace{0.1mm} + CRF~\cite{CRF} \hspace{0.1mm} & \hspace{0.1mm}
+ BI(SLIC)~\cite{BI} \hspace{0.1mm} & \hspace{0.1mm} + BI(ERS) \hspace{0.1mm} & \hspace{0.1mm} + BI(Ours) \hspace{0.1mm}\\
% \midrule
\hline
IoU & 68.9 & 72.7 & 73.5 & 74.0 & \textbf{74.4}\\
\hline
% \bottomrule
\end{tabular}
\vspace{-5mm}
\end{table}

\begin{table}
\centering
\vspace{-5mm}
\caption{\textbf{Superpixels for salient object detection.} We run the SO algorithm with different types of superpixels and evaluate on the ECSSD dataset.}
% \vspace{-2mm}
\label{tab:ecssd}
\begin{tabular}{|c|c|c|c|c|}
\hline
\hspace{6mm} Method \hspace{6mm} & \hspace{6mm} SLIC \hspace{6mm} & \hspace{6mm} SNIC \hspace{6mm} & \hspace{6mm} ERS \hspace{6mm} &\hspace{6mm} Ours \hspace{6mm}\\
\hline
\# of superpixels 200  & 0.1719  & 0.1714 &  0.1686 & \textbf{0.1657} \\
\hline
\# of superpixels 400    & 0.1675 &  0.1654   & 0.1630 & \textbf{0.1616} \\

\hline
% \vspace{-7mm}
\end{tabular}
\vspace{-7mm}
\end{table}

% \begin{table}
%     \centering
%     \caption{\textbf{Superpixels for salient object detection.} We run the SO algorithm with different types of superpixels and evaluate on the ECSSD dataset.}
%     \label{tab:ecssd}
%     \begin{tabular}{lcrr}
%     \hline
%     \multirow{2}{*}{Method} & & \multicolumn{2}{c}{\# of Superpixels} \hspace{-0.5mm}\\
%     & &\hspace{-0.5mm} 200 &\hspace{-0.5mm} 400\\
%     \hline
%     SLIC   &  \multirow{4}*{MAE $\downarrow$} & 0.1719  &  0.1675 \\
%     SNIC    &   & 0.1714    & 0.1654  \\
%     ERS     &   &  0.1686   & 0.1630  \\
%     Ours    &   & \textbf{0.1657} & \textbf{0.1616}  \\
%     \hline
%     \end{tabular}
%     \end{table}

\section{Conclusion}
\vspace{-3mm}

In this paper, we present a superpixel segmentation network SIN which can be integrated into downstream tasks in an end-to-end way.
To extract superpixels, we initialize superpixels and expand pixel-superpixel map multiple times. By dividing an expanding step into a horizontal and a vertical interpolation, we enforce spatial connectivity explicitly.
We utilize multi-layer outputs of a fully convolutional network to predict association scores for interpolations.
To speed up training process, association scores are used to compute loss instead of pixel-superpixel maps.
Owing to our interpolation constrains the number of neighbors of inserted elements, SIN has the fastest speed compared to existing deep learning-based methods.
The high speed of our method ensures it can be integrated into downstream tasks requiring real-time speed.
Our model performs favorably against several existing state-of-the-art superpixel algorithms. SIN can generate more compact superpixels thanks to the design of interpolation, which is important to downstream tasks. What's more, visual results illustrate that our method outperforms when handling fuzzy boundaries. Furthermore, we apply our superpixels in downstream tasks and make progress.
We will integrate SIN into downstream tasks in an end-to-end way in the future and we hope SIN can benefit superpixel-based computer vision tasks.
\vspace{2mm}

\noindent \textbf{Acknowledgements.} This work is supported by the Hubei Provincinal Science and Technology Major Project of China under Grant No. 2020AEA011,
the Key Research \& Developement Plan of Hubei Province of China under Grant No. 2020BAB100, the project of Science,Technology and Innovation Commission of Shenzhen Municipality of China under Grant No. JCYJ20210324120002006 and the Fundamental Research Funds for the Central Universities, HUST: 2020JYCXJJ067.

\bibliographystyle{splncs04}
\bibliography{superpixel}




% LNCS Homepage, \url{http://www.springer.com/lncs}. Last accessed 4
% \end{thebibliography}
\end{document}